\documentclass[hidelinks]{article} 
\usepackage[table,dvipsnames]{xcolor}

\usepackage{colm2024_conference}

\usepackage{microtype}
\usepackage{hyperref}
\usepackage{url}
\usepackage{booktabs}
\usepackage{graphicx}
\usepackage{mdframed}
\usepackage{multirow}

\usepackage{CJKutf8}
\usepackage{wrapfig}
\usepackage{subcaption}
\usepackage{amsmath}
\title{Measuring Taiwanese Mandarin Language Understanding}

\author{Po-Heng Chen\\
National Taiwan University\\
\texttt{r11922044@ntu.edu.tw}
\And
Sijia Cheng\\
National Taiwan University\\
\texttt{r11922184@ntu.edu.tw}
\And
Wei-Lin Chen\\
National Taiwan University\\
\texttt{wlchen@nlg.csie.ntu.edu.tw}
\And
Yen-Ting Lin\\
National Taiwan University\\
\texttt{ytl@ieee.org}
\And
Yun-Nung Chen\\
National Taiwan University\\
\texttt{y.v.chen@ieee.org}
}

\colmfinalcopy 
\begin{document}
\begin{CJK*}{UTF8}{bsmi}

\maketitle

\begin{abstract}
The evaluation of large language models (LLMs) has drawn substantial attention in the field recently.
This work focuses on evaluating LLMs in a Chinese context, specifically, for Traditional Chinese which has been largely underrepresented in existing benchmarks.
We present TMLU, a holistic evaluation suit tailored for assessing the advanced knowledge and reasoning capability in LLMs, under the context of Taiwanese Mandarin.
TMLU consists of an array of 37 subjects across social science, STEM, humanities, Taiwan-specific content, and others, ranging from middle school to professional levels.
In addition, we curate chain-of-thought-like few-shot explanations for each subject to facilitate the evaluation of complex reasoning skills.
To establish a comprehensive baseline, we conduct extensive experiments and analysis on 24 advanced LLMs.
The results suggest that Chinese open-weight models demonstrate inferior performance comparing to multilingual proprietary ones, and open-weight models tailored for Taiwanese Mandarin lag behind the Simplified-Chinese counterparts.
The findings indicate great headrooms for improvement, and emphasize the goal of TMLU to foster the development of localized Taiwanese-Mandarin LLMs.
We release the benchmark and evaluation scripts for the community to promote future research.\footnote{\url{https://huggingface.co/datasets/miulab/tmlu}}

\end{abstract}

{
\definecolor{highschool}{RGB}{182,49,24}
\definecolor{middleschool}{RGB}{71,40,154}
\definecolor{professional}{RGB}{86,114,127}
\definecolor{taiwanspecific}{RGB}{4,77,141}

\begin{figure}[h]
  \begin{center}
  \includegraphics[width=0.8\linewidth]{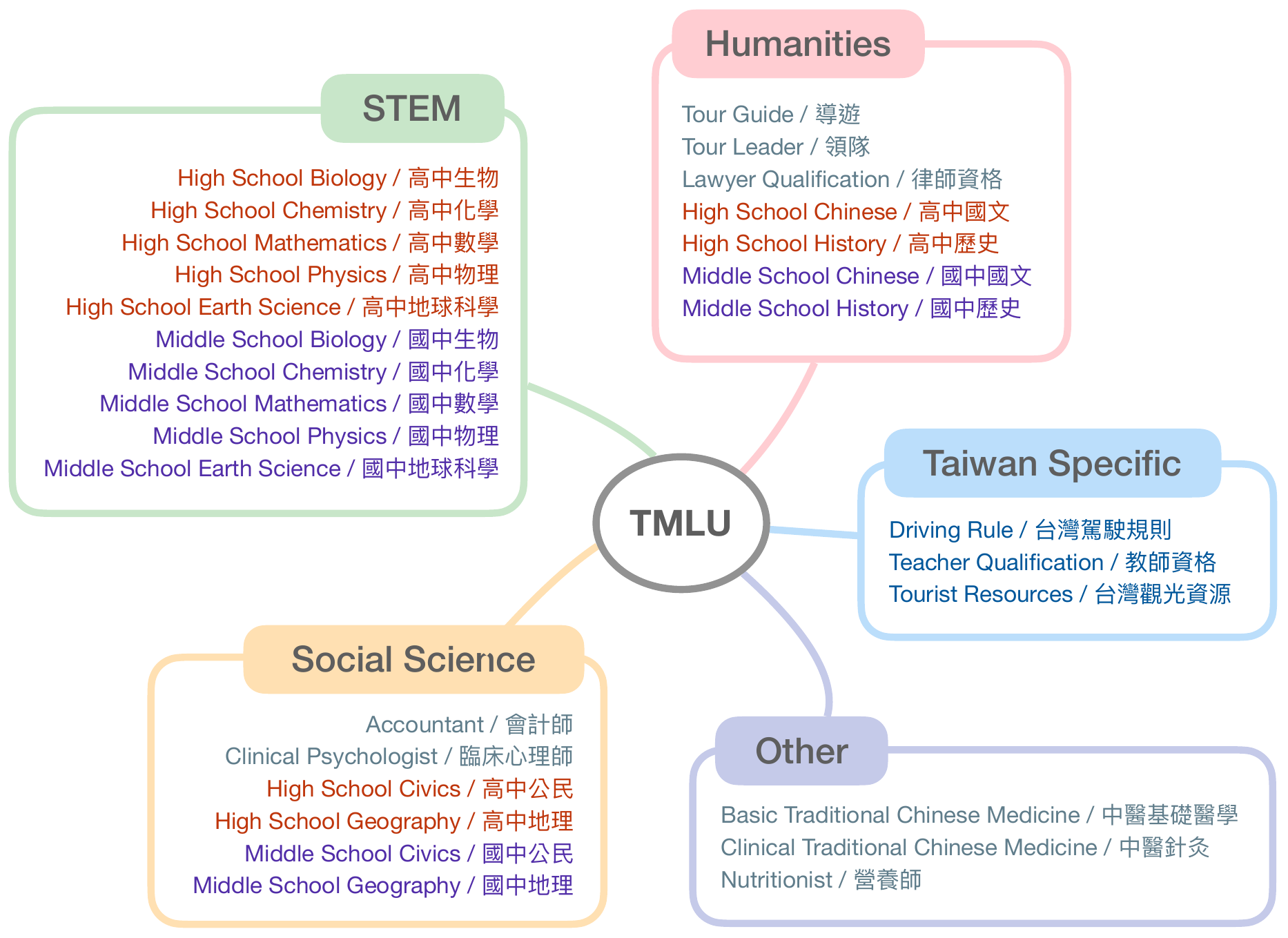}
  \end{center}
  \caption{An overview of our proposed TMLU benchmark.
  TMLU consists of 37 subjects across \textcolor{middleschool}{middle school}, \textcolor{highschool}{high school} and \textcolor{professional}{professional} levels.
  In addition, TMLU includes a set of \textcolor{taiwanspecific}{Taiwan specific questions} in which answers are unique to Taiwanese culture.}
  \label{fig:tmlu_overview}
\end{figure}
}

\section{Introduction}

The emergence of large language models (LLMs) has revolutionized the field of natural language processing (NLP).
As the core that drives the trajectory of AI development, evaluation stands a crucial role in the era of LLMs.
Conventional NLP benchmarks~\citep{wang2018glue,wang2019superglue} have been widely adopted to evaluate natural language understanding (NLU) abilities in LMs.
However, the applicability of these benchmarks has decreased as the ever larger models are demonstrating human-level performance, leaving little headroom for research developments~\citep{hendrycks2020measuring,goyal2022news,liu-etal-2023-g}.

Towards appropriately benchmarking LLMs, probing advanced world knowledge and measuring complex reasoning capabilities are the main focus LLM evalution in recent days~\citep{clark2018think,hendrycks2021measuring,cobbe2021training,wang2023scibench}.
Additionally, evaluation benchmarks for languages beyond English have also been introduced~\citep{li2023cmmlu,huang2023c}, in parallel to the rise of multilingual LLMs~\citep{le2022bloom,muennighoff2022crosslingual,wei2023polylm} and LLMs that are optimized for different regional languages~\citep{JapaneseStableLMBaseAlpha7B,nguyen2023seallms}.

Specifically, a number of benchmarks~\citep{huang2023c,li2023cmmlu,xu2023superclue} have been proposed for assessing Chinese LLMs~\citep{zeng2022glm,yang2023baichuan,bai2023qwen,ai2024yi}.
However, these benchmarks and models are all developed under the context of Simplified Chinese used in Mainland China.
On the other hand, Traditional Chinese, often referred to as Traditional Mandarin, has been notably underrepresented.
Predominantly used in Taiwan, Hong Kong, and Macao, Traditional Mandarin carries different language usage between countries or locations.
Specifically, Taiwanese Mandarin posses linguistic nuances, cultural intricacies, and a different written form which diverge from Chinese used in China and present unique challenges~\citep{lin2023taiwan,wang2021comparison}.
For example, the word ``同志'' refers to ``people with the same interests and ambitions'' in China, while it refers to ``homosexuals'' in Taiwan~\citep{zhou2019study}.
Consequently, the need of building benchmarks organically in Taiwanese Mandarin is necessitated to facilitate the development of localized, Taiwanese-Mandarin LLMs.

In this work, we bridge the gap and present TMLU --- a comprehensive evaluation~suit tailored for assessing advanced knowledge and reasoning capability in LLMs under Taiwanese-Mandarin context.
TMLU consists of a broad spectrum of 37 subjects, spanning social science, STEM, humanities, Taiwan-specific content, and others, from middle school to professional level.
Including high school and college entrance exams, civil service exams, and other localized knowledge in Taiwan.
Furthermore, TMLU includes manually curated, Chain-of-Thought-inspired~\citep{wei2022chain} explanations to facilitate the evaluation of reasoning capability in LLMs.
See~Figure~\ref{fig:tmlu_overview} for an overview of TMLU.

To establish a comprehensive baseline, we evaluate 24 advanced LLMs, including both open-weight and proprietary models, on TMLU.
The experimental results indicate that, in general, proprietary black-box LLMs with multilingual capability outperform open-weight models developed by Chinese communities.
Also, open-weight models tailored for Taiwanese Mandarin demonstrate suboptimal scores comparing to models build predominantly in Simplified-Chinese background.
Overall, the findings suggest that there still exist ample rooms for improvement, underscoring our goal to facilitate the progress of Taiwanese-Mandarin LLMs.
In addition, we present a number of analysis and conduct data contamination test to further validate the reliability and robustness of TMLU.


\section{Related Work} \label{sec:related-work}
Prior to the bloom of LLMs, standard NLP benchmarks such as GLUE \citep{wang2018glue} and SuperGLUE \citep{wang2019superglue} have been widely adopted to evaluate linguistic proficiency and natural language understanding (NLU) abilities in LMs.
However, the practicality of these benchmarks in the era of LLMs have saturated as new models are demonstrating superhuman performance~\citep{goyal2022news,liu-etal-2023-g}.
In light of this, benchmarks which focus on assessing the world knowledge and complex reasoning abilities in LLMs have been introduced recently\citep{hendrycks2020measuring,srivastava2022beyond,hendrycks2021measuring,suzgun2022challenging,liang2022holistic}.
The MMLU benchmark~\citep{hendrycks2020measuring} covers an array of subjects across STEM and social sciences, from elementary to expert levels, for comprehensive multitask evaluation.
The BIG-bench benchmark~\citep{srivastava2022beyond} contains a large collection of diverse tasks that are designed to be beyond the current ability of LLMs.
However, these benchmarks are primarily constructed in English, which limit their usage in assessing LLMs developed in the context of other languages.

To this end, several benchmarks have been proposed to facilitate evaluation of Chinese LLMs.
C-Eval~\citep{huang2023c} presents one of the first comprehensive evaluation suits built to assess LLMs' advanced knowledge and reasoning capabilities in Chinese context, with subjects across social science, STEM, and humanities. 
Another concurrent benchmark, CMMLU~\citep{li2023cmmlu}, shares a similar vision and includes questions relevant to Chinese users' culture and daily life, in addition to standardized examinations.
However, these existing benchmarks are designed in Simplified Chinese, thus, unsuitable for evaluating LLMs developed in the context of Traditional Chinese.

\begin{table}[t]
\small
  \centering
    \begin{tabular}{lccc}
    \toprule
          & \textbf{TC-Eval} & \textbf{TMMLU-plus} & \textbf{TMLU (ours)} \\
    \midrule
    Robustness & medium & low   & high \\
    Localization & low & high  & high \\
    Coverage & low   & high  & medium \\
    Transparency & high  & low   & high \\
    Standardization & high  & medium & high \\
    \bottomrule
    \end{tabular}%
  \caption{A comparison of existing Taiwanese
  Mandarin evaluation benchmarks and our proposed TMLU.
  Notably, TMLU is superior to TC-Eval in the degree of localization, and is more robust to dataset contamination compared to TMMLU-plus.
  We present a detailed discussion in Section~\ref{sec:related-work}.}
  \label{tab:comparison}%
\end{table}%

To the best of our knowledge, only two benchmarks---TC-Eval~\citep{hsu2023advancing} and TMMLU-plus~\citep{tam2024improved}---have been developed towards the goal of evaluating LLMs' capabilities for Traditional Chinese.
TC-Eval consists of a collection of existing datasets, mainly from conventional NLP tasks, \textit{e.g.} reading comprehension, summarization, and sentiment classification~\citep{shao2018drcd,fgc2020stpi,narayan2018don,maas2011learning}, and a newly introduced, MMLU-like multiple-choice dataset, termed TMMLU.
Recently, TMMLU-plus is proposed to focus on the assessment of advance knowledge in Taiwanese-Mandarin LLMs.
Compared to TMMLU, TMMLU-plus is six times larger and contains questions spanning a broader range of subjects.

Our proposed TMLU differs from previous benchmarks in the following ways.
Firstly, all questions within TMLU are originally in Tradition Chinese.
On the other hand, a sizeable portion of TC-Eval is translated from English datasets~\citep{narayan2018don,maas2011learning,srivastava2023beyond}, which misaligns the goal of evaluating LLMs in the Taiwanese-Mandarin environment due to potential western geographical biases.
For example, the XSum dataset~\citep{narayan2018don} adopted by TC-Eval is sourced from BBC articles, which are likely grounded in western economic and social context.
Although TMMLU-plus is tailored specifically for Taiwanese Mandarin, we still observe instances in Simplified Chinese (Figure~\ref{fig:example_instance-simplified-unanswerable}).
Secondly, compared to TMLU, TMMLU-plus exhibits several potential issues that could compromise the reliability and robustness of the benchmark.
In our preliminary investigation, we found that most questions in TMMLU-plus could be found on one single website.\footnote{\url{https://yamol.tw/}}
This greatly exposes its robustness to test data contamination~\citep{sainz2023chatgpt,sainz-etal-2023-nlp,golchin2023time,shi2023detecting} if the website is scrapped for pre-training.\footnote{We observed the presence of the website's (\textit{i.e.}, \url{https://yamol.tw/}) content in mC4~\citep{xue2021mt5}, a widely adopted corpus for pre-training LMs.}
To minimize the risk of test data contamination, we follow~\citet{huang2023c} by collecting data sourced in PDF and Microsoft Word documents from the internet, instead of plain texts directly from the web.
Moreover, TMMLU-plus contains unanswerable questions that require information (\textit{e.g.}, images or tables) not accompanied
in the dataset (Figure~\ref{fig:example_instance-simplified-unanswerable}).

Lastly, TMLU provides two distinct features not available in TC-Eval and TMMLU-plus:
(1) manual-constructed few-shot CoT demonstrations to elicit reasoning;
(2) standardized mathematical expressions (\textit{e.g}., symbols and equations) in \LaTeX{} format for better clarity (Figure~\ref{fig:example_instance-compare_format}).
In sum, our proposed TMLU is better for the community to objectively evaluate the model capability due to its better transparency\footnote{The source of data collection is not disclosed in the TMMLU-plus paper.}, localization, and robustness.









\section{TMLU Benchmark}
\subsection{Overview}
The TMLU benchmark is an evaluation suit tailored for assessing advanced knowledge and reasoning capability in LLMs under Taiwanese Mandarin, in the format of multiple-choice questions.
TMLU contains a wide range of subjects spanning social science, STEM, humanities, Taiwan-specific content, and others, across middle school to professional levels.
The goal of TMLU is to provide an evaluation suit, which is easy-to-use and accessible, for developers to gauge how their models would likely perform in the real-world Taiwanese Mandarin context.

\begin{table}[h]
\footnotesize
  \centering
    \begin{tabular}{lrr}
    \toprule
          & \multicolumn{1}{l}{\textbf{\# Subjects}} & \multicolumn{1}{l}{\textbf{\# Instance (\%)}} \\
    \midrule
    \multicolumn{3}{c}{\textit{group by discipline}} \\
    Social Science &   8    &  589 (19.76$\%$)\\
    STEM  &   14    &  468 (15.70$\%$)\\
    Humanities &   9    &  1,009 (33.85$\%$)\\
    Taiwan Specific &  3     &  557 (18.69$\%$)\\
    Others &    3   &  358 (12.01$\%$)\\
    \midrule
    \multicolumn{3}{c}{\textit{group by level}} \\
    Middle School &    9   &  434 (14.56$\%$)\\
    High School &   17    &  875 (29.35$\%$)\\
    Professional &  11    &  1,672 (56.09$\%$)\\
    \midrule
    Total &   37   & 2,981 (100.00$\%$) \\
    \bottomrule
    \end{tabular}%
    \caption{Statistics of our proposed TMLU benchmark grouped by different categories of discipline and difficulty levels.}
  \label{tab:tmlu-breakdown}%
\end{table}%


\subsection{Data Source} \label{subsec:data-source}
One of the main concerns of NLP evaluation in the age of LLMs is test data contamination~\citep{sainz2023chatgpt,sainz-etal-2023-nlp,golchin2023time,shi2023detecting}, sometimes referred to as data leakage.
That is, the models are (pre-)trained on data that belongs to the test split of a benchmark and subsequently evaluate on the same test split, making the evaluation results questionable.
As these is generally no guarantee of building contamination-free evaluation benchmark, especially given how little information about the training data are disclosed by proprietary models, mitigating the risk of contamination is equally important.

To this end, we follow the design principle of~\citealt{huang2023c} and collect data that is sourced in the format of PDF or Microsoft Word documents, which are files downloaded from the website instead of readily available within website plain text.
Specifically, we source our data from the following five websites from Taiwan:\footnote{The sourced website links are provided in Table~\ref{tab:data-source-website}.}
\begin{enumerate}
    \item \textit{Comprehensive Assessment Program for Junior High School Students} (CAP)
    -- the standardized exam for Taiwanese 9-th grade students before they go to high schools or vocational schools.
    
    \item \textit{General Scholastic Ability Test} (GSAT)
    -- the Taiwanese university entrance exam, focusing on fundamental knowledge and skills from the required materials in the first two years of students' high school studies.
    
    \item \textit{Advanced Subjects Test} (AST)
    -- the other part of the Taiwanese university entrance exam, focusing on advanced knowledge of individual subjects to assess students' comprehension, reasoning, and analysis abilities.
    The scope covers all materials in the high school studies.
    
    \item \textit{Ministry of Examination Exams} (MOEX)
    -- the exams from Ministry of Examination of Taiwan consists of two main categories, civil service exams and professional and technical exams, spanning a wide range of field such as law, finance, medical, and pharmacy.
    
    \item \textit{Highway Bureau Questions Bank} (HB)
    -- the theory test questions from the Highway Bureau of Taiwan, including questions such as car and motorcycle driving regulations for the driving license test.
\end{enumerate}

\subsection{Data Processing}
All of our data is in the format of PDF or Microsoft Word document initially and processed into text format subsequently, by manual annotation with the aided of OCR toolkits.\footnote{\url{https://mathpix.com/}}
Mathematical expressions are standardized into \LaTeX{} format for STEM subjects that involve substantial symbols and equations, following prior works~\citep{hendrycks2021measuring,huang2023c}.
The produced \LaTeX{} formulas are compiled and verified by annotators to ensure their correctness.
In addition, all instances are validated by human for quality assurance to exclude unsuitable instances (\textit{e.g.}, unanswerable questions that require accompanied images or tables which are not presented in the text body).

Our final resulting benchmark, TMLU, consists of a collection of 2,981 multiple-choice questions across 37 subjects, covering difficulty levels from middle school, high school, to professional.
We categorize the subjects into five disciplines -- social science, STEM, humanities, Taiwan specific, and other.
Each subject is further divided into a development set consisting of five instances and a test set consisting of the remaining instances.
Statistics of TMLU grouped by difficulty disciplines and levels are provided in Table~\ref{tab:tmlu-breakdown}.

\subsection{Explanation Curation} \label{subsec:explanation-curation}
To facilitate the evaluation of reasoning capability in Taiwanese-Mandarin LLMs, we manually construct Chain-of-Thought-like explanations for each development set instances.
Specifically, the main source of our explanations (55.68\%) are curated by leveraging the textbook website.\footnote{\url{https://www.ehanlin.com.tw/app/index.html}}
It provides corresponding textbook-level explanations for CAP, GSAT, and AST exam questions included in TMLU, in the format of PDF documents.
Other sources for explanation curation include
\textit{(1)} question-and-answer websites or forum (26.49\%);
\textit{(2)} GPT-4-preview (3.78\%), where explanations are adopted only if the predicted answer is correct;
\textit{(3)} fully handcrafted (4.86\%), which is composed by thoroughly searching the Web;
\textit{(4)} others (9.19\%), where the exact source is either ambiguous or untraceable.


\begin{figure}[t]
  \begin{center}
  \includegraphics[width=0.75\linewidth]{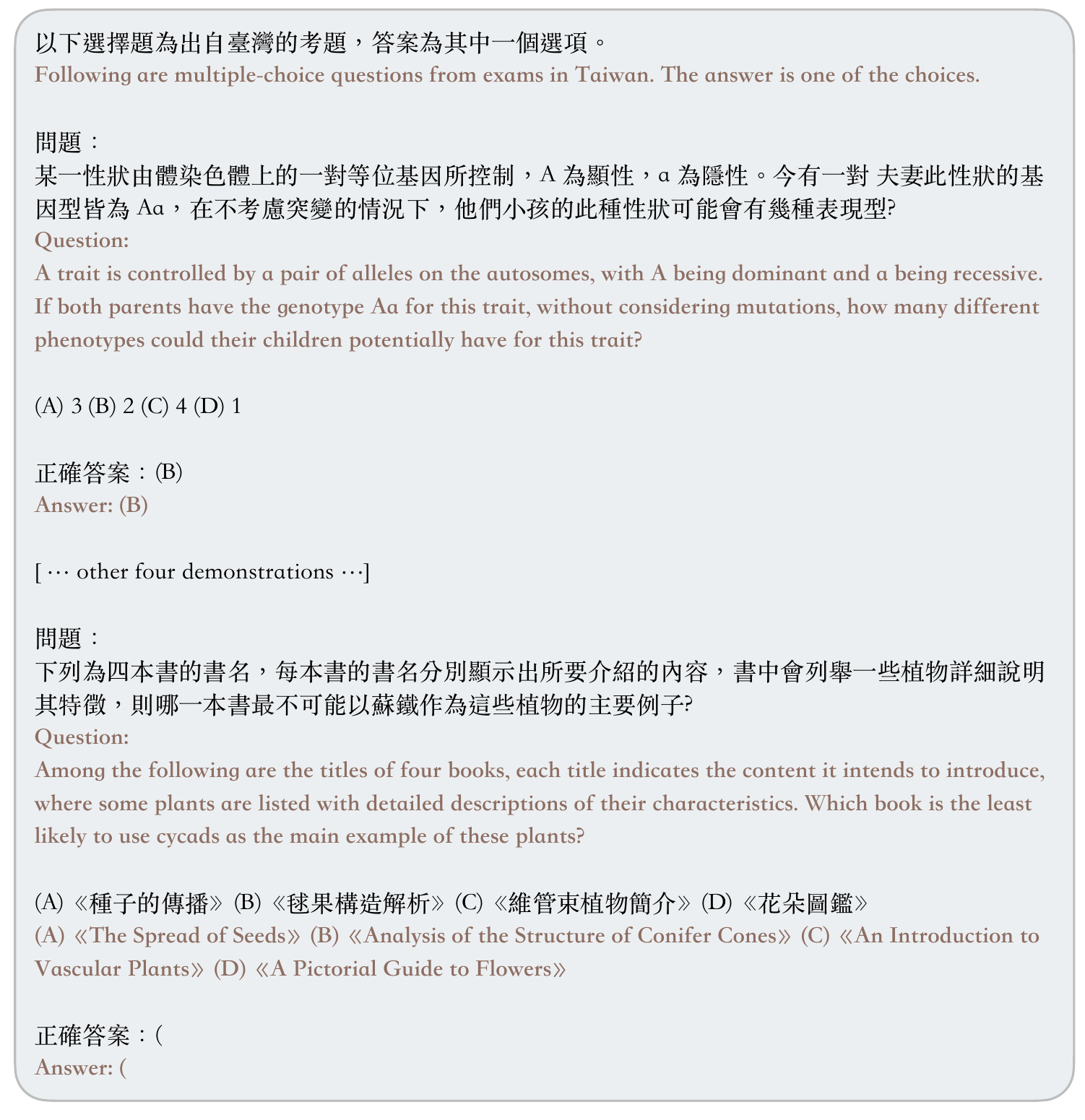}
  \end{center}
  \caption{An example prompt for few-shot direct answer evaluation on TMLU.}
  \label{fig:tmlu-prompt-example-direct}
\end{figure}

\section{Experiments}

We conduct evaluation on an array of advanced LLMs on TMLU, establishing a baseline for research and community reference.
Following we describe the experimental details and results.

\subsection{Experimental Setups}
\paragraph{Models.}
We comprehensively evaluation TMLU on 24 different LLMs capable of processing Traditional-Chinese content, varying in size and developer organizations.
Concretely, we conduct experiment on 6 closed-source, proprietary models via APIs, including models from OpenAI,\footnote{We adopt gpt-4-0125-preview and gpt-3.5-turbo-1106.} Anthropic, and Google.
For open-weight models, we focus on models developed for Chinese, English, and multilingual purposes.
This included Taiwan-LLM~\citep{lin2023taiwan} and Breeze~\citep{hsu2024breeze7b} for Taiwanese Mandarin, as well as Yi~\citep{ai2024yi} and Qwen~\citep{bai2023qwen} for Simplified Chinese. 
For a complete list of the models evaluated, please refer to Table~\ref{tab:results-all}.

\paragraph{Few-shot Evaluation.}
The evaluation is conducted in few-shot with setups: direct answer and CoT.
For each subject, the five instances from the development split are utilized as demonstrations (\textit{i.e.}, five shots).
The adoption of few-shot evaluation is consistent with benchmarks such as C-Eval~\citep{huang2023c} and MMLU~\citep{hendrycks2020measuring}, and the convention in prior works~\citep{touvron2023llama,team2023gemini,achiam2023gpt} for measuring LLM performance.
The setting is believe to capture the inherent, underlying potential of LLMs more robustly and extrapolate to downstream task adaptation~\citep{huang2023c}.

The evaluation prompt consists of the following components:
the problem definition, a set of few-shot examples, and the input instance to query. 
For instruction-tuned (chat-oriented) models, we apply the accompanied chat templates if available.
Compared with direct answer, CoT prompting includes an explanation (\textit{i.e.}, rationale) triggered with ``\textit{Let's think step by step}", before the final answer prediction.
Example prompts with TMLU instances are provided in Figure~\ref{fig:tmlu-prompt-example-direct} (direct answer) and Figure~\ref{fig:tmlu-prompt-example-cot} (CoT).

\paragraph{Answer Extraction.}
As show by~\citealp{fourrier2023s}, different flavors of implementation in extracting predictions from models bring nontrivial impact to the evaluation results.
To ensure a fair evaluation and address variations in model accessibility and prompting settings, we adopt two methods for answer extraction:
(1) \textit{Likelihood-based} method.
The method aligns with the original implementation in MMLU ~\citep{hendrycks2021measuring}, which involves examining a set of candidate answer symbols (\textit{i.e.,} \{``\textit{A}", ``\textit{B}", ``\textit{C}", ``\textit{D}"\}) and selecting the symbol with the highest assigned token probability as the model's prediction.
(2) \textit{Generation-based} method.
When direct determination of candidate token probabilities is impractical, we opt for the first token generated greedily that can be interpreted as an option code to derive the answer prediction.

For proprietary models, as access to the probability of each candidate answer symbol is restricted, we utilize the generation-based method for both direct and CoT prompting scenarios.
In the case of open-source models, we employ the likelihood-based method for direct prompting. However, when employing CoT prompting, where the model must first generate an explanation for the queried instance, we are constrained to using the generation-based method.  In instances of CoT prompting, we adjust the number of shots based on the model context length to avoid failure to produce the answer.

{
\begin{table}[ht]    
\begin{subtable}{\linewidth}
\scriptsize
\begin{center}
\begin{tabular}{lccccc|c}
\toprule
\multicolumn{1}{c}{\bf Model} &\multicolumn{1}{c}{\bf Social Sci.} &\multicolumn{1}{c}{\bf STEM} &\multicolumn{1}{c}{\bf Humanities} &\multicolumn{1}{c}{\bf Taiwan Spe.} &\multicolumn{1}{c|}{\bf Other} &\multicolumn{1}{c}{\bf Average} \\
\midrule
\cellcolor{Goldenrod!40}Claude-3-Opus$^\pi$              &\textbf{83.06} &\textbf{64.32} &\textbf{75.93} &\textbf{90.41} &54.23 &\textbf{73.59} \\
\cellcolor{Goldenrod!40}GPT-4-turbo$^\pi$                &78.14 &58.79 &72.30 &86.90 &55.98 &70.42 \\
\cellcolor{NavyBlue!30}Yi-34B-Chat                   &73.77 &47.99 &73.55 &83.95 &\textbf{62.97} &68.45 \\
\cellcolor{NavyBlue!30}Qwen-14B-Chat                 &64.85 &45.98 &64.83 &79.15 &54.23 &61.81 \\
\cellcolor{Goldenrod!40}Gemini-Pro$^\pi$                &69.22 &41.96 &65.87 &81.55 &48.40 &61.40 \\
\cellcolor{NavyBlue!30}Qwen-7B-Chat                  &57.74 &39.20 &56.54 &73.99 &46.65 &54.82 \\
\cellcolor{Goldenrod!40}Claude-Instant-1.2$^\pi$        &60.84 &36.43 &56.33 &75.46 &42.86 &54.38 \\
\cellcolor{Salmon!50}Breeze-7B-Instruct              &57.19 &36.68 &49.79 &73.80 &39.07 &51.31 \\
\cellcolor{Goldenrod!40}GPT-3.5$^\pi$                    &54.64 &35.43 &46.37 &72.51 &37.90 &49.37 \\
\cellcolor{Goldenrod!40}Mixtral-8x7B-Instruct      &51.55 &33.17 &46.47 &66.05 &34.99 &46.44 \\
\cellcolor{Goldenrod!40}chatglm3-6b                   &51.55 &33.17 &46.47 &66.05 &34.99 &46.44 \\
\cellcolor{Goldenrod!40}Claude-2.0$^\pi$                 &40.62 &30.90 &41.08 &73.06 &37.32 &44.60 \\
\cellcolor{Goldenrod!40}Mistral-7B-Instruct        &46.08 &30.65 &42.43 &67.71 &34.40 &44.26 \\
\cellcolor{Salmon!50}Taiwan-LLM-13B-Chat             &44.08 &27.89 &41.80 &64.39 &36.15 &42.86 \\
\cellcolor{Goldenrod!40}Baichuan2-13B-Chat         &46.08 &31.91 &42.84 &56.64 &28.57 &41.21 \\
\cellcolor{Salmon!50}Taiwan-LLM-7B-Chat              &35.15 &21.86 &32.99 &64.02 &32.94 &37.39 \\
\cellcolor{YellowGreen!35}falcon-40b-instruct          &36.07 &21.36 &36.41 &51.48 &32.07 &35.48 \\
\cellcolor{YellowGreen!35}OLMo-7B-Instruct             &35.70 &25.38 &29.88 &50.37 &31.78 &34.62 \\
\cellcolor{YellowGreen!35}Llama-2-13b-Chat             &36.07 &27.14 &35.27 &47.05 &26.24 &34.35 \\
\cellcolor{NavyBlue!30}Yi-6B-Chat                    &35.52 &26.38 &37.66 &34.50 &29.74 &32.76 \\
\cellcolor{NavyBlue!30}Qwen-0.5B-Chat                &30.78 &24.12 &29.15 &45.57 &28.57 &31.64 \\
\cellcolor{YellowGreen!35}Llama-2-7b-chat              &29.33 &23.37 &27.28 &32.84 &27.41 &28.04 \\
\cellcolor{YellowGreen!35}Falcon-7b-instruct           &22.59 &22.86 &23.76 &34.32 &24.78 &25.66 \\
\cellcolor{YellowGreen!35}Gemma-7b-it                  &27.32 &20.10 &24.07 &32.66 &23.91 &25.61 \\
\bottomrule
\end{tabular}
\end{center}
\setlength{\fboxsep}{1pt}
\caption{The five-shot direct-answer accuracy results. (\%)}
\label{tab:results-direct}
\end{subtable}

\qquad

\begin{subtable}{\linewidth}
\scriptsize
\begin{center}
\begin{tabular}{lccccc|c}
\toprule
\multicolumn{1}{c}{\bf Model} &\multicolumn{1}{c}{\bf Social Sci.} &\multicolumn{1}{c}{\bf STEM} &\multicolumn{1}{c}{\bf Humanities} &\multicolumn{1}{c}{\bf Taiwan Spe.} &\multicolumn{1}{c|}{\bf Other} &\multicolumn{1}{c}{\bf Average} \\
\midrule
\cellcolor{Goldenrod!40}Claude-3-Opus$^\pi$              &\textbf{82.15} &\textbf{74.37} &\textbf{73.44} &\textbf{89.67} &\textbf{53.06} &\textbf{74.54} \\
\cellcolor{Goldenrod!40}GPT-4-turbo$^\pi$                &75.77 &74.12 &70.12 &86.35 &51.31 &71.54 \\
\cellcolor{Goldenrod!40}Gemini-Pro$^\pi$                 &64.66 &47.24 &59.54 &82.66 &45.77 &59.97 \\
\cellcolor{Goldenrod!40}Claude-2.0$^\pi$                 &66.67 &54.02 &58.51 &80.81 &39.65 &59.93 \\
\cellcolor{Goldenrod!40}Claude-Instant-1.2$^\pi$         &61.57 &44.47 &56.43 &75.83 &37.03 &55.07 \\
\cellcolor{Goldenrod!40}GPT-3.5$^\pi$                    &59.20 &45.48 &49.90 &75.65 &41.69 &54.38 \\
\cellcolor{NavyBlue!30}Yi-6B-Chat                    &46.45 &34.17 &46.27 &70.85 &39.07 &47.36 \\
\cellcolor{NavyBlue!30}Yi-34B-Chat                    &49.54 &18.84 &30.39 &79.34 &33.53 &42.33 \\
\cellcolor{Salmon!50}Breeze-7B-Instruct              &48.63 &22.61 &30.19 &74.35 &27.11 &40.58 \\
\cellcolor{Goldenrod!40}Mixtral-7B-Instruct      &42.81 &23.87 &37.14 &59.23 &30.61 &38.73 \\
\cellcolor{NavyBlue!30}Qwen-7B-Chat                 &39.71 &18.09 &29.25 &64.94 &30.03 &36.41 \\
\cellcolor{Goldenrod!30}chatglm3-6b                   &37.34 &21.11 &32.47 &63.10 &23.91 &35.58 \\
\cellcolor{Salmon!50}Taiwan-LLM-7B-Chat              &26.05 &23.62 &19.81 &56.27 &24.49 &30.05 \\
\cellcolor{Goldenrod!40}Baichuan2-13B-Chat         &29.51 &19.85 &27.07 &30.26 &27.11 &26.76 \\
\cellcolor{Salmon!50}Taiwan-LLM-13B-Chat             &26.41 &12.56 &17.63 &48.15 &14.29 &23.81 \\
\cellcolor{NavyBlue!30}Qwen-0.5B-Chat                &22.04 &18.59 &20.02 &34.69 &18.08 &22.68 \\
\bottomrule
\end{tabular}
\end{center}
\setlength{\fboxsep}{1pt}
\caption{The five-shot CoT accuracy results. (\%)}
\label{tab:results-cot}
\end{subtable}

\setlength{\fboxsep}{1pt}
\caption{The five-shot accuracy (\%) results on TMLU for direct-answer and CoT prompting.
The results are sorted in descending order.
The colors indicate the language for which the models are tailored.
\colorbox{Goldenrod!35}{Yellow} denotes Multilingual;
\colorbox{NavyBlue!30}{Blue} denotes Simplified Chinese;
\colorbox{Salmon!40}{Red} denotes Taiwanese Mandarin;
\colorbox{YellowGreen!40}{Green} denotes English.
In addition, proprietary models are marked with $\pi$.
Detailed comparison results are provided at Table~\ref{tab:direct_vs_cot}.}
\label{tab:results-all}
\end{table}
}

\subsection{Results}
The experimental results are presented in Table~\ref{tab:results-direct} and Table~\ref{tab:results-cot} for direct answer and CoT prompting, respectively.
In general, flagship proprietary models demonstrate performance superior to open-weight counterparts, and models trained on Chinese data (\textit{i.e.}, Simplified Chinese, Taiwanese Mandarin, and Multilingual models) exhibit substantial improvement over models focusing on English context.
For direct answer prompting, the Breeze model -- tailored for Taiwanese Mandarin with only 7 billions parameters -- outperforms GPT-3.5.
The top-performing open-weight model Yi, with 34 billions parameters, outperforms proprietary models such as Gemini-pro and Claude-instant, and is comparable to GPT-4.

In CoT prompting, proprietary models outperform the best open-source model by a significant margin of 7\%$\sim$27\%.
The results suggest an ample opportunity for enhancing reasoning capabilities via CoT prompting in open-source models.
We provide more analysis between direct answer and CoT prompting in Section~\ref{sec:analysis}.

In sum, the established baseline suggests that TMLU offers a holistic ground, with headroom for improvements, for evaluating LLMs in the context of Taiwanese Mandarin.
Furthermore, the exhibited performance of Taiwanese-Mandarin LLMs still lack behind the Simplified-Chinese counterparts, substantiating the goal of TMLU to foster the development of localized, Taiwanese-Mandarin LLMs in the future.

\section{Analysis} \label{sec:analysis}

\subsection{Robustness to Test Data Contamination}

To further investigate the possibility of test data contamination mentioned in Section~\ref{sec:related-work} and~\ref{subsec:data-source}, we employ \textsc{Min-k}\% \textsc{Prob}~\citep{shi2023detecting}, a reference-free method for detecting pre-training data from LLMs.
Based on the hypothesis that an example seen by the model before (\textit{i.e.}, during pre-training) is less likely to include words with low probability (\textit{i.e.}, high negative log-likelihood), \textsc{Min-k}\% \textsc{Prob} selects a set of $k$\% of tokens from the input text with minimum token probability and average their log-likelihood as an indicator of whether the input text is in the pre-training data.
Specifically, given an input example $x = (x_1, x_2, ..., x_n)$,
\begin{equation*}
    \textsc{Min-k}\%\,\,\textsc{Prob} (x) = \frac{1}{\lvert \textrm{\textrm{Min-K(x)}} \rvert} \sum_{x_i \in \textrm{Min-K(x)}} -\log p(x_i|x_1,x_2,..x_{i-1})
\end{equation*}

where Min-K(x) is the set of $k$\% of tokens in $x$ with the minimum token probability.
The lower the \textsc{Min-k}\% \textsc{Prob} is, the more likely that the input example $x$ has been seen during pre-training.




\begin{figure}[h]
  \begin{center}
  \includegraphics[width=\linewidth]{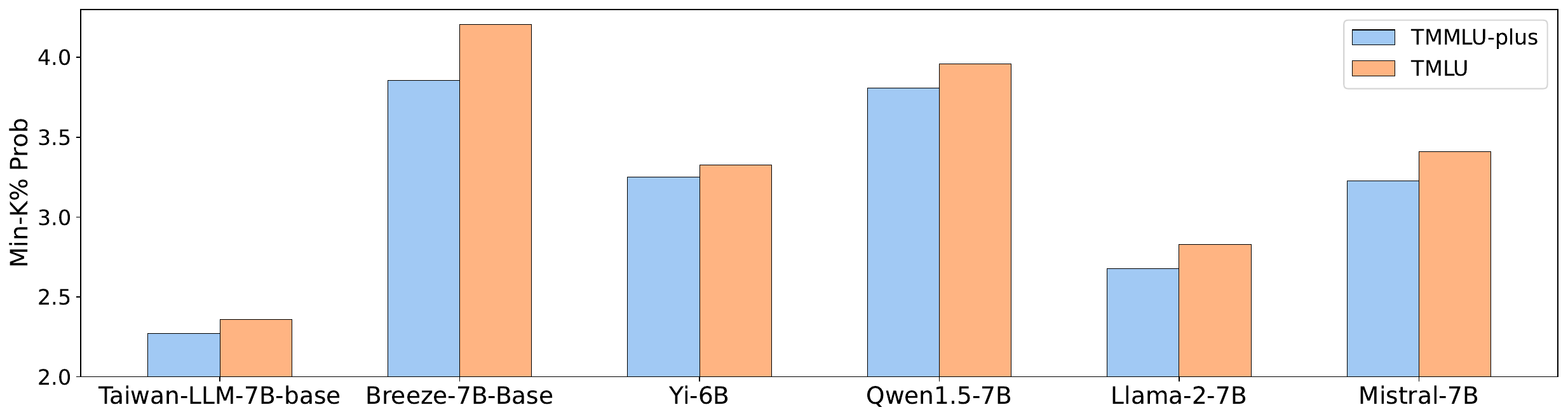}
  \end{center}
  \caption{The \textsc{Min-k}\% \textsc{Prob}~\citep{shi2023detecting} of six base models on TMMLU-plus~\citep{tam2024improved} and our dataset TMLU.
  The lower the \textsc{Min-k}\% \textsc{Prob} is, the more likely the input instances of the datasets are in the model's pre-training data.}
  \label{fig:pre-training-detection}
\end{figure}

We apply \textsc{Min-k}\% \textsc{Prob} on 2000 sampled instances from TMLU and TMMLU-plus with 6 different base models.
We present results in Table~\ref{fig:pre-training-detection} and discuss more implementation details in Appendix~\ref{app:data-contamination-detail}.
As shown, TMLU achieves higher \textsc{Min-k}\% \textsc{Prob} than TMMLU-plus across all tested models, which implies that TMLU is less likely to contain data that has been seen during pre-trained.
In addition, the finding also aligns with our hypothesis that TMMLU-plus might be collected by crawling plain text from the web, and further validates the effectiveness of our design principle (Section~\ref{subsec:data-source}) that sourcing data from downloaded documents could reduce the risk of data contamination.

\subsection{Comparison Between Direct Answer and CoT Prompting}

\begin{figure}[h]
  \begin{center}
  \begin{minipage}[b]{0.4\textwidth}
  \includegraphics[width=\linewidth]{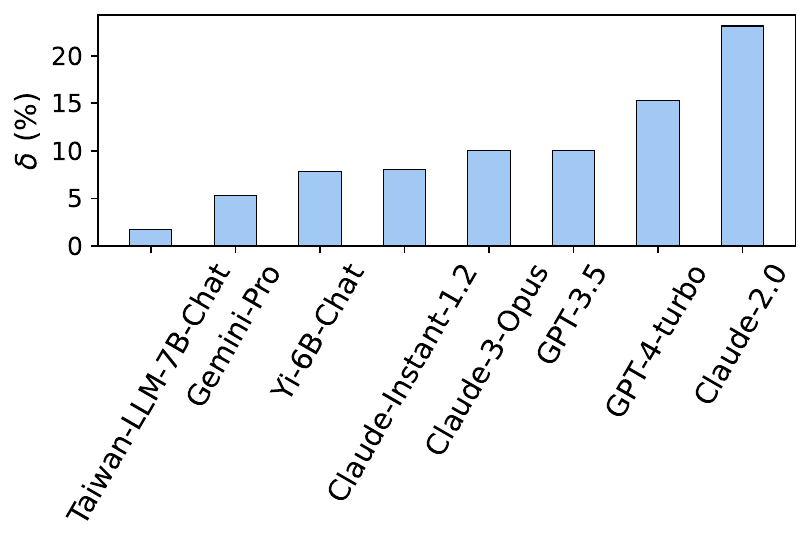}
  \caption{Performance difference ($\delta$) between direct answer and CoT prompting on stem subjects.
  Only models exhibiting improvements ($\delta>0$) are presented.}
  \label{fig:stem-cot-vs-direct}
  \end{minipage}
  \hfill
  \begin{minipage}[b]{0.55\textwidth}
  \includegraphics[width=\textwidth]{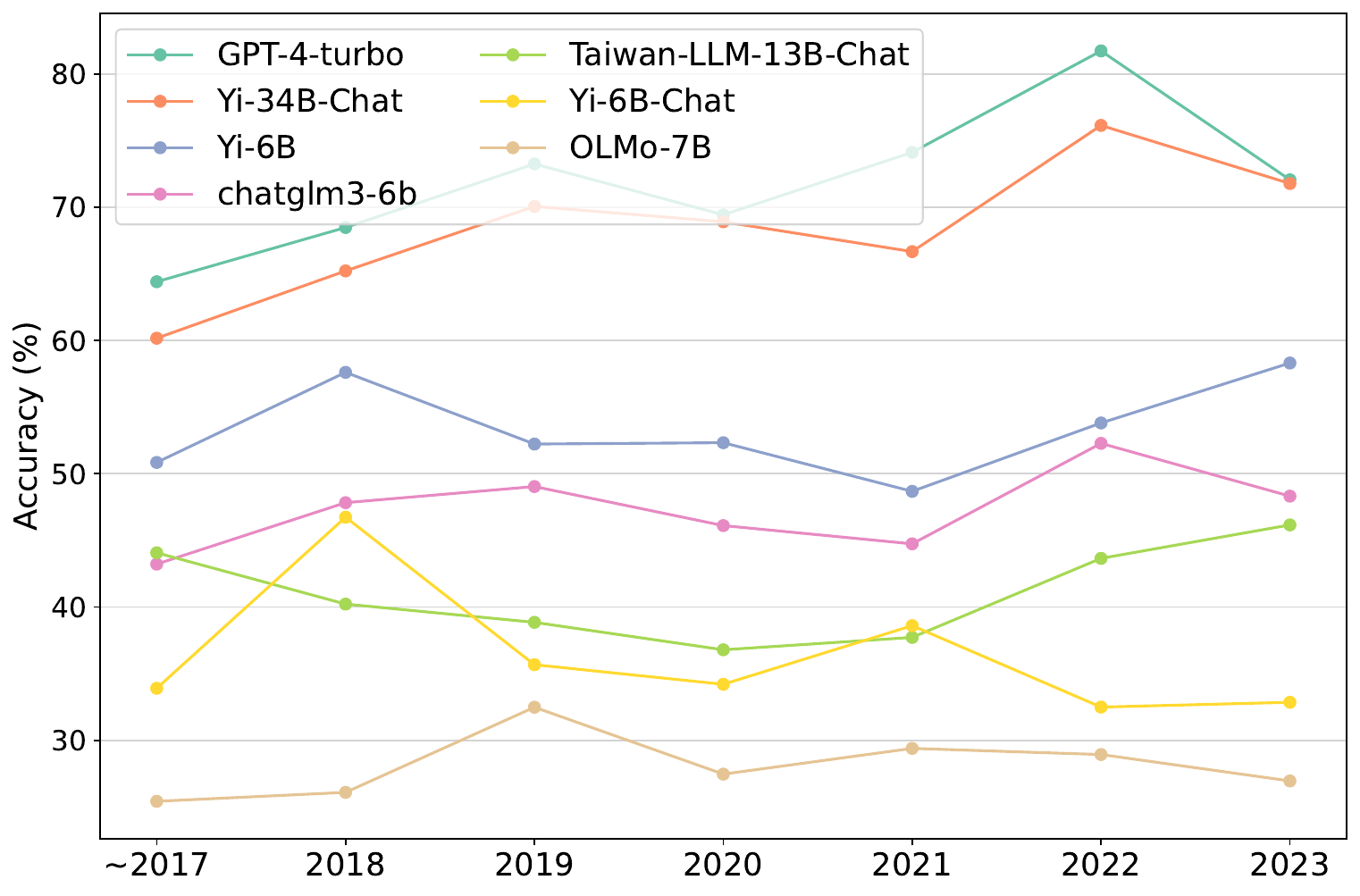}
  \caption{Average accuracy of models on questions of different years. The accuracy is calculated by averaging across the number of questions. Full results are provided at Table~\ref{tab:year_comparison}.}
  \label{fig:compare-time}
  \end{minipage}
  \end{center}
\end{figure}

To further investigate the potential capability of LLMs, we leverage our curated explanations and employ CoT prompting.
CoT prompting elicits step-by-step reasoning chains towards answer derivations, which has been shown to improve tasks that require complex, multi-hop reasoning significantly.
We present results comparing CoT with standard answer-only prompting in Figure~\ref{fig:stem-cot-vs-direct} for STEM subjects.
Full results are reported in Table~\ref{tab:direct_vs_cot}.
As shown, models that are able to benefit from CoT are mainly large, proprietary ones, exemplified by GPT-4 (15\%) and Claude-2 (23\%).
The finding may be attribute to the emergent nature of reasoning capability in LLMs.
Interestingly, Taiwan-LLM-7B-Chat is the only model exhibiting improvement with CoT at its parametric scale.



\subsection{Comparison of Model Performance Across the Temporal Dimension}


Here we investigate the performance of models on the temporal dimension.
We group questions from different years as subsets and evaluate the average accuracy, as presented in Figure~\ref{fig:compare-time}.
Questions from before 2017 (2013$\sim$2016) are group with 2017 as one subset.
The results show that, for most models, no consistent trends or behaviors are exhibited.
Yet, for GPT-4-turbo and Yi-34B-Chat, the scores imply a rising trend where models perform better on questions from more recent years.
Future endeavors could offer a more precise implication and rigorous explanation for the results.

\section{Conclusion}
In this work, we introduce TMLU -- an evaluation suite designed to assess the advanced knowledge and reasoning abilities of LLMs in the context of Taiwanese Mandarin.
Experimental results and analysis suggest great opportunities for future developers, in particular, for open-source models.
Besides addressing the scarcity of benchmark for Traditional Chinese community, we envision TMLU to serve as a grounded testbad, fostering development of localized Taiwanese-Mandarin LLMs.

\bibliography{colm2024_conference}
\bibliographystyle{colm2024_conference}

\appendix

\section{Limitations and Future Work}

While TMLU provides a comprehensive evaluation of language models' understanding capabilities in Taiwanese Mandarin, it is important to acknowledge its limitations, particularly in terms of its generalizability to generation tasks. Our benchmark, which focuses on multiple-choice question answering, may not fully capture the performance of language models in generation scenarios such as text completion, summarization, and dialogue systems.

To investigate this issue, we conduct a preliminary experiment by calculating the perplexity (PPL) of generated answers for a few models with a fixed vocabulary size of 32,000 tokens. The resulting PPL values were: Taiwan-LLM-7B-Chat: $16.5$, Breeze-7B-instruct: $8836.1$, Yi-6B-Chat: $20887.0$, and Llama-7B-Chat: $2038633.8$. Interestingly, models trained exclusively on Taiwanese Mandarin, such as Taiwan-LLM, have a smaller entropy in their logits compared to models trained on multilingual data, suggesting more consistent performance in generating fluent Taiwanese Mandarin text.

Therefore, while TMLU provides valuable insights into the understanding capabilities of language models in Taiwanese Mandarin, it should not be considered a sole indicator of their performance in generation tasks. Future research should focus on developing generation-oriented benchmarks and exploring the relationship between understanding and generation abilities to gain a more holistic view of the performance of language models in the context of Taiwanese Mandarin.

\section{Implementation Details of Data Contamination Analysis} \label{app:data-contamination-detail}
We construct the input example $x$ for \textsc{Min-k}\% \textsc{Prob} by concatenating the question and the corresponding choices -- consistent to the actual scenario where the model would likely be queried -- for each sampled instances from TMLU and TMMLU-plus.
To mitigate the effect of input example length when computing \textsc{Min-k}\% \textsc{Prob}, as discussed in the \textsc{Min-k}\% \textsc{Prob} paper, we filtered out the longest examples within each subjet subset of our dataset, as the average text length of TMLU is significantly longer than that of TMMLU-plus.
This step is taken to ensure a fair and comparable analysis.

\begin{table}[t]
\scriptsize
  \centering
    \begin{tabular}{ll}
    \toprule
    \textbf{Data source} & \textbf{Website} \\
    \midrule
    CAP   & \url{https://cap.rcpet.edu.tw/examination.html} \\
    GSAT  & \url{https://www.ceec.edu.tw/xmfile?xsmsid=0J052424829869345634} \\
    AST   & \url{https://www.ceec.edu.tw/xmfile?xsmsid=0J052427633128416650} \\
    MOEX  & \url{https://wwwq.moex.gov.tw/exam/wFrmExamQandASearch.aspx} \\
    HB    & \url{https://www.thb.gov.tw/News\_Download.aspx?n=284\&sms=12823} \\
    \bottomrule
    \end{tabular}%
  \caption{The sourced websites of our data collection.}
  \label{tab:data-source-website}%
\end{table}%

\begin{figure}[h]
     \centering
     \begin{subfigure}[b]{\textwidth}
         \centering
         \includegraphics[width=\textwidth]{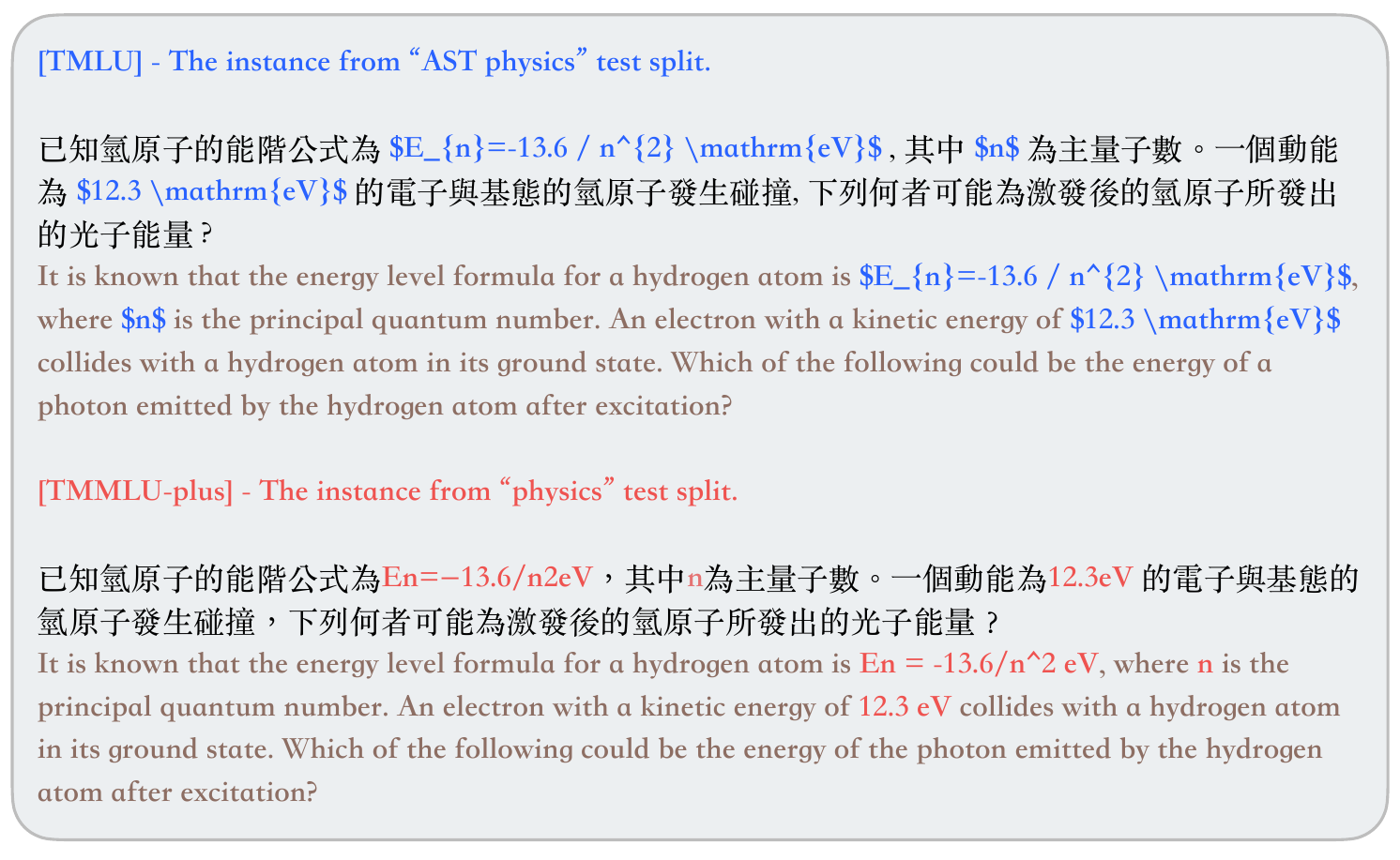}
         \caption{}
         \label{fig:example_instance-compare_format}
     \end{subfigure}
     \hfill
     \begin{subfigure}[b]{\textwidth}
         \centering
         \includegraphics[width=\textwidth]{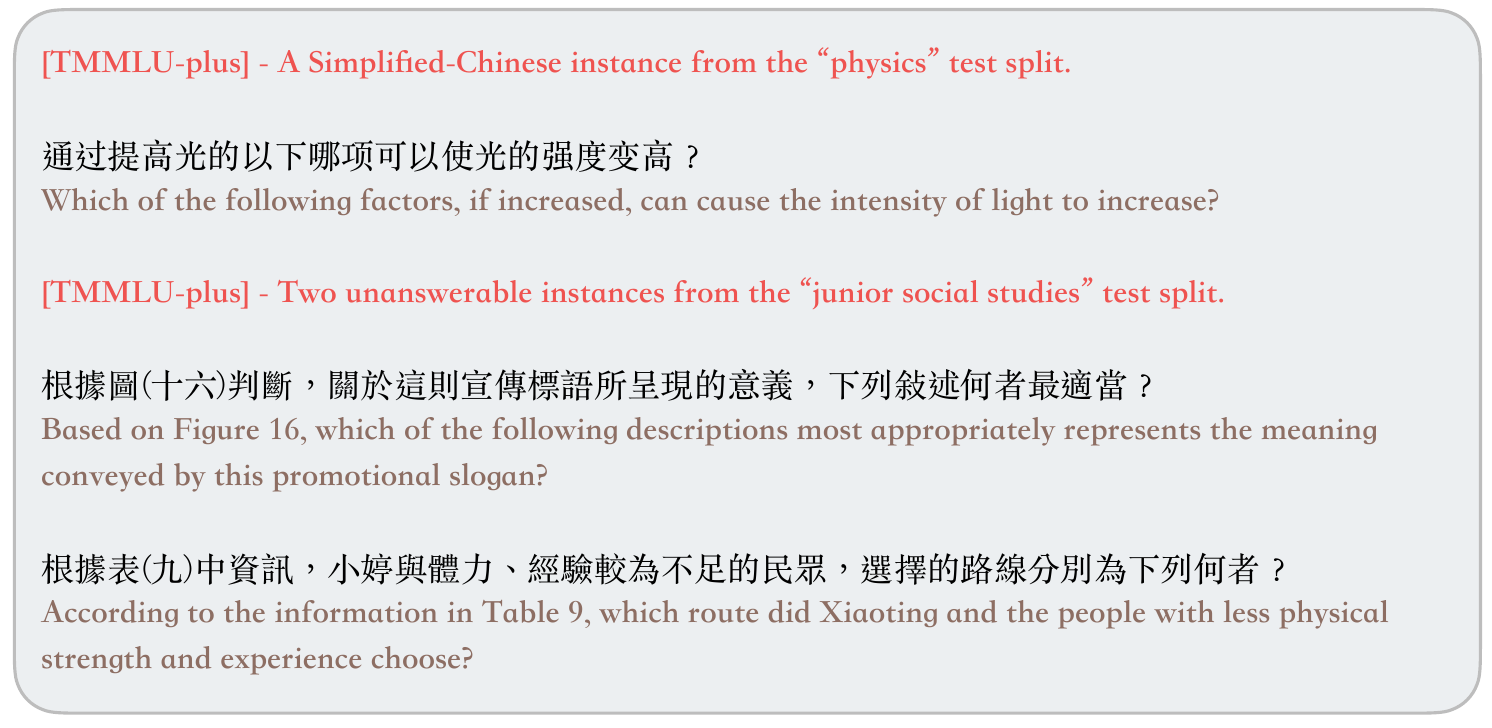}
         \caption{}
         \label{fig:example_instance-simplified-unanswerable}
     \end{subfigure}
        \caption{Example instances from TMLU and TMMLU-plus datastes.
        Here choices are omitted for brevity.
        Figure~\ref{fig:example_instance-compare_format} shows a same instance from TMLU (AST\_physics) and TMMLU-plus (physics\_test), where symbols are formatted in latex in TMLU but are not formatted in TMMLU-plus, effecting the clarity.
        Figure~\ref{fig:example_instance-simplified-unanswerable} shows instances from TMMLU-plus (junior\_chemistry\_test and junior\_social\_studies\_test) which are in Simplified Chinese or unanswerable since it requires information from an image/table that is not accompanied in the question.}
        \label{fig:example_instance}
\end{figure}

\begin{figure}[b]
  \begin{center}
  \includegraphics[width=\linewidth]{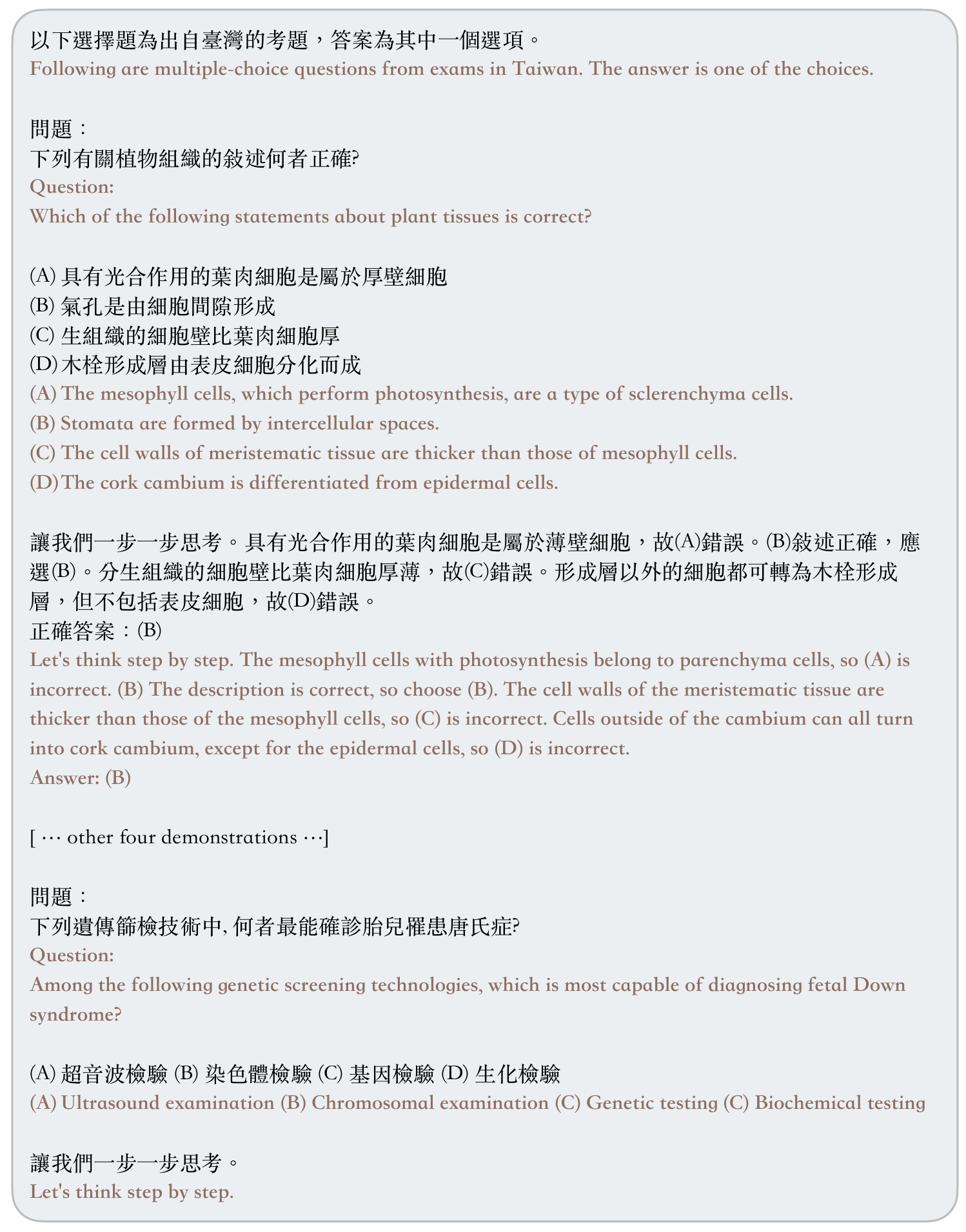}
  \end{center}
  \caption{An example prompt for few-shot CoT evaluation on TMLU.}
  \label{fig:tmlu-prompt-example-cot}
\end{figure}

\begin{table}[t]
\scriptsize
\label{summary-table}
\begin{center}
\begin{tabular}{clrr}
\toprule
\multicolumn{1}{c}{\bf Supercategory} & \multicolumn{1}{c}{\bf Subjects}  &\multicolumn{1}{c}{\bf \# Questions} & \bf \# Choices \\
\midrule
Social Science & AST civics (分科測驗公民)          &57  & 4\\
& AST geography (分科測驗地理)                      &58  & 4\\
& CAP civics (會考公民)                            &73  & 4\\
& CAP geography (會考地理)                         &45  & 4\\
& GSAT civics (學測公民)                           &73  & 4\\
& GSAT geography (學測地理)                        &49  & 4\\
& MOEX Accountant (會計師)                              &117 & 4\\
& MOEX Clinical psychologist (臨床心理師)               &117 & 4\\
\midrule
STEM & AST biology (分科測驗生物)                 &40  & 4\\
& AST chemistry (分科測驗化學)                      &34  & 5\\
& AST mathematics (分科測驗數學)                    &25  & 5 \\
& AST physics (分科測驗物理)                        &43  & 5\\
& CAP biology (會考生物)                           &27  & 4\\
& CAP chemistry (會考化學)                         &27  & 4\\
& CAP earth science (會考地球科學)                  &15 & 4\\
& CAP mathematics (會考數學)                       &115 & 4\\
& CAP physics (會考物理)                           &15  & 4\\
& GSAT biology (學測生物)                          &21  & 5\\
& GSAT chemistry (學測化學)                        &29  & 5\\
& GSAT earth science (學測地球科學)                 &24  & 5\\
& GSAT mathematics (學測數學)                      &29  & 5\\
& GSAT physics (學測物理)                          &24  & 5\\
\midrule
Humanities & AST Chinese (分科測驗國文)               &131 & 4 \\
& AST history (分科測驗歷史)                        &56  & 4\\
& CAP Chinese (會考國文)                        &61 & 4\\
& CAP history (會考歷史)                           &56 & 4 \\
& GSAT Chinese (學測國文)                          &97  & 4\\
& GSAT history (學測歷史)                          &85  & 4\\
& MOEX Tour guide (導遊)                               &99  & 4\\
& MOEX Tour leader (領隊)                              &145  & 4\\
& MOEX Lawyer qualification (律師資格)                  &279 & 4\\
\midrule
Taiwan Specific & HB Driving Rule (台灣駕駛規則)      &432  & 4\\
& MOEX Teacher qualification (教師資格)                 &75  & 4\\
& MOEX Taiwan tourist resources (台灣觀光資源)          &50  & 4\\
\midrule
Others & MOEX Basic Traditional Chinese Medicine (中醫基礎醫學) &159 & 4\\
& MOEX Clinical Traditional Chinese Medicine (中醫針灸) &79  & 4\\
& MOEX Nutritionist (營養師)                            &120 & 4\\
\bottomrule
\end{tabular}
\end{center}
\caption{Summary of all 37 subjects and corresponding numbers of choices.}
\end{table}

\begin{table}[ht]
\scriptsize
\begin{center}
\begin{tabular}{l|ccccccc|c}
\toprule
Model & $\sim$2017 & 2018 & 2019 & 2020 & 2021 & 2022 & 2023  & Avg. \\ 
\midrule
\# of questions & 118 & 92 & 157 & 193 & 228 & 197 & 1811  &  \\ 
\midrule
GPT-4-turbo & 64.41 & 68.48 & 73.25 & 69.43 & 74.12 & 81.73 & 72.06  & 72.35  \\ 
Yi-34B-Chat & 60.17 & 65.22 & 70.06 & 68.91 & 66.67 & 76.14 & 71.78  & 70.67  \\ 
Gemini-pro & 56.78 & 65.22 & 63.06 & 61.14 & 61.40 & 67.51 & 64.77  & 64.02  \\ 
Qwen-7B-Chat & 48.31 & 56.52 & 59.87 & 51.30 & 54.39 & 57.36 & 57.43  & 56.47  \\ 
Claude-instant-1.2 & 49.15 & 56.52 & 55.41 & 48.70 & 52.19 & 56.85 & 58.31  & 56.44  \\ 
Yi-6B & 50.85 & 57.61 & 52.23 & 52.33 & 48.68 & 53.81 & 58.31  & 56.12  \\ 
GPT-3.5 & 51.69 & 44.57 & 52.23 & 44.04 & 41.67 & 55.33 & 51.79  & 50.46  \\ 
Mixtral-8x7B-Instruct & 43.22 & 51.09 & 52.23 & 43.01 & 42.98 & 54.31 & 50.19  & 49.25  \\ 
chatglm3-6b & 43.22 & 47.83 & 49.04 & 46.11 & 44.74 & 52.28 & 48.32  & 47.96  \\ 
Mistral-7B-Instruct & 39.83 & 47.83 & 43.31 & 42.49 & 28.95 & 38.58 & 48.92  & 45.39  \\ 
Claude-2.0 & 38.98 & 46.74 & 40.76 & 38.34 & 41.23 & 42.64 & 47.54  & 45.28  \\ 
Taiwan-LLM-13B-Chat & 44.07 & 40.22 & 38.85 & 36.79 & 37.72 & 43.65 & 46.16  & 43.96 \\
Baichuan2-13B-Chat & 37.29 & 42.39 & 50.96 & 38.34 & 40.35 & 54.31 & 42.08 & 42.85  \\ 
Taiwan-LLM-7B-Chat & 27.97 & 29.35 & 32.48 & 30.57 & 27.63 & 33.50 & 41.91 & 37.84  \\ 
falcon-40b-instruct & 30.51 & 33.70 & 34.39 & 35.23 & 31.58 & 36.55 & 38.10 & 36.59  \\ 
Llama-2-13b-Chat & 40.68 & 34.78 & 32.48 & 29.53 & 32.89 & 40.61 & 35.78 & 35.44  \\ 
Yi-6B-Chat & 33.90 & 46.74 & 35.67 & 34.20 & 38.60 & 32.49 & 32.85 & 34.05  \\ 
Qwen-0.5B-Chat & 35.59 & 27.17 & 26.11 & 27.98 & 26.75 & 30.46 & 33.57 & 31.87  \\ 
Llama-2-7b-Chat & 32.20 & 32.61 & 28.03 & 22.80 & 30.26 & 24.87 & 28.44 & 28.22  \\ 
OLMo-7B & 25.42 & 26.09 & 32.48 & 27.46 & 29.39 & 28.93 & 26.95 & 27.54  \\ 
gemma-7b-it & 20.34 & 17.39 & 24.84 & 21.76 & 25.00 & 28.93 & 26.84 & 25.79  \\ 
falcon-7b-instruct & 14.41 & 20.65 & 23.57 & 20.21 & 26.75 & 30.96 & 26.56 & 25.57  \\
\midrule
Total & 41.34 & 44.09 & 44.34 & 41.37 & 41.73 & 46.81 & 46.70 & 45.45 \\
\bottomrule
\end{tabular}
\end{center}
\caption{Performance comparison on questions from different years. The accuracy is calculated by averaging across the number of questions.}
\label{tab:year_comparison}
\end{table}

\begin{table}[ht]
\scriptsize
\begin{center}
\begin{tabular}{l|c}
\toprule
Year & \# of questions \\
\midrule
2013 & 10 \\
2014 & 15\\
2015 & 27\\
2016 & 27\\
2017 & 39\\
2018 & 92\\
2019 & 157\\
2020 & 193\\
2021 & 228\\
2022 & 197\\
2023 & 1811\\
\bottomrule
\end{tabular}
\end{center}
\caption{Number of questions from different years.}
\label{tab:number of questions in different years}
\end{table}

\begin{table}[ht]
\footnotesize
\begin{center}
\resizebox{\textwidth}{!}{\begin{tabular}{l|cc|cc|cc|cc|cc|cc}
\toprule
\multirow{2}{*}{\bf Model} &\multicolumn{2}{c|}{\bf Social Science} &\multicolumn{2}{c|}{\bf STEM} &\multicolumn{2}{c|}{\bf Humanities} &\multicolumn{2}{c|}{\bf Taiwan Specific} &\multicolumn{2}{c|}{\bf Other} &\multicolumn{2}{c}{\bf Avg.} \\
&Base &Chat &Base &Chat &Base &Chat &Base &Chat &Base &Chat &Base &Chat \\
\midrule
Yi-34B                   &77.96 &73.77 &51.51 &47.99 &73.76 &73.55 &86.35 &83.95 &69.68 &62.97 &71.85 &68.45 \\
Qwen-14B              &66.12 &64.85 &47.49 &45.98 &65.77 &64.83 &78.97 &79.15 &57.14 &54.23 &63.10 &61.81 \\
Yi-6B                    &60.29 &35.52 &40.20 &26.38 &60.27 &37.66 &78.04 &34.50 &50.44 &29.74 &57.85 &32.76 \\
Qwen-7B               &60.47 &57.74 &39.45 &39.20 &60.37 &56.54 &75.46 &73.99 &52.19 &46.65 &57.59 &54.82 \\
Baichuan2-13B       &55.56 &46.08 &31.41 &31.91 &53.63 &42.84 &72.69 &56.64 &46.36 &28.57 &51.93 &41.21 \\
Mistral-7B          &51.91 &46.08 &31.41 &30.65 &43.46 &42.43 &68.08 &67.71 &35.28 &34.40 &46.03 &44.26 \\
Taiwan-LLM-13B &42.08 &44.08 &25.38 &27.89 &44.09 &41.80 &67.90 &64.39 &35.28 &36.15 &42.94 &42.86 \\
Llama-2-13b           &44.44 &36.07 &31.66 &27.14 &39.32 &35.27 &62.36 &47.05 &32.36 &26.24 &42.03 &34.35 \\
Qwen-0.5B             &36.61 &30.78 &26.88 &24.12 &35.58 &29.15 &52.58 &45.57 &26.82 &28.57 &35.70 &31.64 \\
Taiwan-LLM-7B  &34.79 &35.15 &24.62 &21.86 &29.67 &32.99 &57.56 &64.02 &27.11 &32.94 &34.75 &37.39 \\
Llama-2-7b            &31.69 &29.33 &26.88 &23.37 &29.98 &27.28 &55.54 &32.84 &23.32 &27.41 &33.48 &28.04 \\
falcon-7b                &23.50 &22.59 &25.13 &22.86 &28.53 &23.76 &32.29 &34.32 &25.36 &24.78 &26.96 &25.66 \\
OLMo-7B                  &23.50 &35.70 &23.37 &25.38 &26.66 &29.88 &34.50 &50.37 &24.78 &31.78 &26.56 &34.62 \\
gemma-7b                 &25.14 &27.32 &23.62 &20.10 &24.79 &24.07 &29.34 &32.66 &25.07 &23.91 &25.59 &25.61 \\
\bottomrule
\end{tabular}}
\end{center}
\caption{Performance comparison between base and instruction-tuned models.}
\label{tab:base_vs_chat}
\end{table}

\begin{table}[ht]
\footnotesize
\begin{center}
\resizebox{\textwidth}{!}{\begin{tabular}{l|cc|cc|cc|cc|cc|cc}
\toprule
\multirow{2}{*}{\bf Model}  &\multicolumn{2}{c|}{\bf Social Science} &\multicolumn{2}{c|}{\bf STEM} &\multicolumn{2}{c|}{\bf Humanities} &\multicolumn{2}{c|}{\bf Taiwan Specific} &\multicolumn{2}{c|}{\bf Other} &\multicolumn{2}{c}{\bf Avg.} \\
&Direct &CoT &Direct &CoT &Direct &CoT &Direct &CoT &Direct &CoT &Direct &CoT \\
\midrule
Claude-3-Opus   &83.06 &82.15 &64.32 &74.37 &75.93 &73.44 &90.41 &89.67 &54.23 &53.06 &73.59 &74.54 \\
GPT-4-turbo       &78.14 &75.77 &58.79 &74.12 &72.30 &70.12 &86.90 &86.35 &55.98 &51.31 &70.42 &71.54 \\
Yi-34B-Chat &73.77 &49.54 &47.99 &18.84 &73.55 &30.39 &83.95 &79.34 &62.97 &33.53 &68.45 &42.33 \\
Qwen-14B-Chat &64.85 &55.19 &45.98 &25.63 &64.83 &45.85 &79.15 &78.04 &54.23 &48.98 &61.81 &50.74 \\
Gemini-Pro               &69.22 &64.66 &41.96 &47.24 &65.87 &59.54 &81.55 &82.66 &48.40 &45.77 &61.40 &59.97 \\
Qwen-7B-Chat &57.74 &39.71 &39.20 &18.09 &56.54 &29.25 &73.99 &64.94 &46.65 &30.03 &54.82 &36.41 \\
Claude-Instant-1.2       &60.84 &61.57 &36.43 &44.47 &56.33 &56.43 &75.46 &75.83 &42.86 &37.03 &54.38 &55.07 \\
Breeze-7B-Instruct &57.19 &48.63 &36.68 &22.61 &49.79 &30.19 &73.80 &74.35 &39.07 &27.11 &51.31 &40.58 \\
GPT-3.5       &54.64 &59.20 &35.43 &45.48 &46.37 &49.90 &72.51 &75.65 &37.90 &41.69 &49.37 &54.38 \\
ChatGLM3-6B &51.55 &37.34 &33.17 &21.11 &46.47 &32.47 &66.05 &63.10 &34.99 &23.91 &46.44 &35.58 \\
Claude-2.0               &40.62 &66.67 &30.90 &54.02 &41.08 &58.51 &73.06 &80.81 &37.32 &39.65 &44.60 &59.93 \\
Mistral-7B-Instruct &46.08 &42.81 &30.65 &23.87 &42.43 &37.14 &67.71 &59.23 &34.40 &30.61 &44.26 &38.73 \\
Taiwan-LLM-13B-Chat &44.08 &26.41 &27.89 &12.56 &41.80 &17.63 &64.39 &48.15 &36.15 &14.29 &42.86 &23.81 \\
Baichuan2-13B-Chat &46.08 &29.51 &31.91 &19.85 &42.84 &27.07 &56.64 &30.26 &28.57 &27.11 &41.21 &26.76 \\
Taiwan-LLM-7B-Chat &35.15 &26.05 &21.86 &23.62 &32.99 &19.81 &64.02 &56.27 &32.94 &24.49 &37.39 &30.05 \\
Llama-2-13v-Chat &36.07 &23.13 &27.14 &20.10 &35.27 &28.53 &47.05 &39.30 &26.24 &26.24 &34.35 &27.46 \\
Yi-6B-Chat  &35.52 &46.45 &26.38 &34.17 &37.66 &46.27 &34.50 &70.85 &29.74 &39.07 &32.76 &47.36 \\
Qwen-0.5B-Chat &30.78 &22.04 &24.12 &18.59 &29.15 &20.02 &45.57 &34.69 &28.57 &18.08 &31.64 &22.68 \\
Llama-2-7B-Chat &29.33 &9.47 &23.37 &7.04 &27.28 &7.68 &32.84 &1.66 &27.41 &9.33 &28.04 &7.03 \\
\bottomrule
\end{tabular}}
\end{center}
\caption{Performance comparison between direct and CoT prompting.}
\label{tab:direct_vs_cot}
\end{table}

\end{CJK*}
\end{document}